\documentclass[11pt]{article}
\pdfoutput=1

\usepackage{microtype}
\usepackage{geometry}
\usepackage{times}
\usepackage{url}

\usepackage{coling2020}

\usepackage{booktabs}
\usepackage{hyperref}
\usepackage{latexsym}
\usepackage{upgreek}
\usepackage{amsmath} 
\usepackage{amsthm}
\usepackage{array}
\usepackage{multirow}

\DeclareMathOperator*{\argmax}{argmax}
\newcolumntype{L}[1]{>{\raggedright\arraybackslash}p{#1}}

\colingfinalcopy % Uncomment this line for all SemEval submissions

\title{CIRCE at SemEval-2020 Task 1: \\ Ensembling Context-Free and Context-Dependent Word Representations}

\author{Martin P{\"o}msl\thanks{\quad Work done during internship at Cogent Labs Inc.} \\
	Osnabr{\"u}ck University \\
	{\tt mpoemsl@uos.de} \\ \\\And
	Roman Lyapin \\
	Cogent Labs Inc. \\
	{\tt rlyapin@cogent.co.jp} \\}

\date{}

\begin{document}
	
\maketitle

\begin{abstract}
  This paper describes the winning contribution to SemEval-2020 Task 1: Unsupervised Lexical Semantic Change Detection (Subtask 2) handed in by team UG Student Intern. We present an ensemble model that makes predictions based on context-free and context-dependent word representations. The key findings are that (1) context-free word representations are a powerful and robust baseline, (2) a sentence classification objective can be used to obtain useful context-dependent word representations, and (3) combining those representations increases performance on some datasets while decreasing performance on others.
\end{abstract}

\section{Introduction}
\label{sec:introduction}

SemEval-2020 Task 1 poses an evaluation framework for unsupervised Lexical Semantic Change Detection (LSCD). Its two subtasks operate on a set of non-parallel corpus pairs from two different time periods and are evaluated against human annotations for semantic change of a subset of words. Subtask 1 requires binary classification of whether or not the meaning of the words has changed. Subtask 2 requires ranking the words by degree of lexical semantic change and is evaluated in Spearman's rank-order correlation coefficient $\uprho$ \cite{Schlechtweg2020TaskDescript}. This paper primarily addresses Subtask 2.

One of the most successful methods for predicting the degree of semantic change of words is comparing context-free (static) semantic vector spaces. Such models seperately induce word vectors for all words in two corpora (e.g. with Word2Vec \cite{Mikolov2013SkipGram}), align the resulting vector spaces and take the distance of the word vectors as a measure of semantic change \cite{Schlechtweg2019Woc}.

With recent advances in language model pretraining, it is now possible to extract context-dependent (contextualized) word representations for each use of a word in the two corpora by using a language model (e.g. BERT \cite{Devlin2019Bert}) as feature extractor. The distance between these context-dependent word representations can then be taken as a measure of semantic change \cite{Giulianelli2019ContextChange}.

In this paper, we present a model based on context-free word representations, a model based on context-dependent word representations and an ensemble model that combines their predictions. We obtain context-free representations by following the methodology of the best model reported by \newcite{Schlechtweg2019Woc}, aligning SGNS vectors. For context-dependent representations, we finetune BERT with a sentence classification objective - predicting the time period of sentences - and extract internal representations for all words from the finetuned BERT model. We show that this classification finetuning can both produce useful word representations and provide an indicator for how to parameterize the ensemble.

In the evaluation phase of SemEval-2020 Task 1, the context-free model ranked first out of 128 contributions to Subtask 2. The ensemble model performed better in one language, but significantly worse in another, causing it to be ranked fifth among all contributions. The context-dependent model ranked 69th, suggesting that it can sometimes add information to the context-free model, but is largely inadequate on its own. In the results section, we analyze the submission experiments and show that the usefulness of the context-dependent representations is linked to the BERT classification accuracy. Code and predictions for all models are available publicly at \url{https://github.com/mpoemsl/circe}.

\blfootnote{
	\hspace{-0.65cm} This work is licensed under a Creative Commons 
	Attribution 4.0 International License. \\ License details: \url{http://creativecommons.org/licenses/by/4.0/}.
}

\section{Related Work}
\label{sec:related-work}

\subsection{LSCD with Context-Free Word Representations}

Approaches that rely on the comparison of context-free word representations in diachronic corpora have a long history in lexical semantic change detection. The word vectors are usually either explicitly derived from co-occurrence statistics or implicitly through the use of neural methods \cite{Tahmasebi2018Survey}. Context-free models often follow a three-step scheme: Representing words in semantic vectors, aligning the resulting vector spaces and comparing relevant word vectors.

\newcite{Kim2014StaticTemp} use Skip-Gram, the neural word embedding method introduced by \newcite{Mikolov2013SkipGram} to represent words over multiple diachronic corpora and compare the representations. A similar method that includes an alignment step was used by \newcite{Hamilton2016Alignment}.

\subsection{LSCD with Context-Dependent Word Representations}

Context-dependent word representations assign a semantic vector to each word-use within the context of its sentence, rather than to each unique word. One way to get context-dependent word representations is using a pretrained neural language model as a feature extractor \cite{Peters2018DeepContext}. Such representations have proven useful for a wide range of NLP tasks \cite{Liu2019ContextRep} and are becoming increasingly popular for LSCD. Like context-free models, context-dependent models usually follow a three-step scheme: Extracting semantic vectors for each use of the relevant words, clustering the resulting semantic vectors and comparing the mean clusters with a distance metric.

\newcite{Hu2019ContextSense} use context-dependent representations derived from a neural language model as the basis for word sense tracking. \newcite{Giulianelli2019ContextChange} clusters the resulting word-use representations into usage type distributions, which can then be compared.

\section{Available Datasets}
\label{sec:available-datasets}

A complete dataset for lexical semantic change ranking as defined by \newcite{Schlechtweg2019Woc} consists of two lemmatized corpora from different time periods ($t_{1}$ and $t_{2}$) and a corresponding testset. This testset contains gold ranks for a subset of words as annotated by human experts. Predictions are made for all target words on the basis of the two corpora and evaluated against the true ranks in the testset.

In the development experiments, we validate our approach by following \newcite{Schlechtweg2019Woc} in evaluating our models on the diachronic testset DURel \cite{Schlechtweg2018Durel} (de-durel) and the synchronic testset SURel \cite{Hatty2019Surel} (de-surel) in combination with the corresponding corpora. It should be noted that de-surel presents domain-specific rather than time-specific meaning differences, but since diachronic and synchronic lexical semantic change detection are closely related, it is still a useful development dataset.

In the submission experiments, we compute our submission for the evaluation phase based on the diachronic datasets provided in SemEval-2020 Task 1 \cite{Schlechtweg2020TaskDescript}, which consist of an English (en-semeval), German (de-semeval), Latin (ln-semeval) and Swedish (sw-semeval) dataset. Results are averaged over submission datasets. An overview of all datasets can be found in Table \ref{tab:data-table}.

\begin{table}[b!]
	\begin{tabular}{@{}llp{0.10\textwidth}*{5}{L{\dimexpr0.15\textwidth-2\tabcolsep\relax}}@{}}
		\toprule
		& & Language & Total Sentences & Annotated Words & Time Period $t_{1}$ & Time Period $t_{2}$ \\
		\midrule
		\multirow{2}{*}{Development} & de-durel & German & $ 3.2 \cdot 10^{6} $ & 19 & 1750 - 1799 & 1850 - 1899 \\
		& de-surel & German & $ 0.9 \cdot 10^{6} $ & 21 & - & - \\
		\midrule
		\multirow{4}{*}{Submission} & en-semeval & English & $ 0.6 \cdot 10^{6} $ & 37 & 1810 - 1860 & 1960 - 2010 \\
		& de-semeval & German & $ 6.1 \cdot 10^{6} $ & 48 & 1800 - 1899 & 1946 - 1990 \\
		& ln-semeval & Latin & $ 0.6 \cdot 10^{6} $ & 40 & -200\, - 0 & 0 \enspace \, \, \,- 2000 \\
		& sw-semeval & Swedish & $ 8.5 \cdot 10^{6} $ & 31 & 1790 - 1830 & 1895 - 1903 \\
		\bottomrule
	\end{tabular}
	
	\caption{Properties of datasets used in development and submission experiments.}
	\label{tab:data-table}
\end{table}

\newpage

\section{System Overview}
\label{sec:system-overview}

We present three models for SemEval-2020 Task 1 Subtask 2: A context-free model, a context-dependent model and the ensemble model CIRCE, which stands for Classification-Informed Representation Comparison Ensemble. For Subtask 1, we binarize the CIRCE rank predictions by naively assuming that the upper half of ranks has changed while the lower half has not.

\subsection{Context-Free Model}

The context-free model is structured analogously to the best performing model reported by \newcite{Schlechtweg2019Woc}. We adopt the use of word vectors generated by Skip-Gram with Negative Sampling (SGNS) \cite{Mikolov2013SGNS} as context-free word representations. Similarly, we follow their use of orthogonal Procrustes analysis  \cite{Schonemann1966OrthogonalProcrustes} to align the embeddings. 

However, we diverge from the best performing model reported by \newcite{Schlechtweg2019Woc} in that we employ Euclidean distance rather than cosine distance to compare the aligned representations, since this metric achieved more robust results in development experiments.

\subsection{Context-Dependent Model}

The context-dependent model follows a similar outline to the one described in \newcite{Giulianelli2019ContextChange}, with a few key changes. We adopt the use of context-dependent representations derived from the masked language model BERT \cite{Devlin2019Bert}. We also recognize the need for domain-adaptive finetuning of the pretrained BERT model as described by \newcite{Han2019Finetuning}. 

However, instead of the standard language modelling objective, we use a sentence time classification objective. This is motivated by the assumption that a successful time classifier for sentences must learn time-specific word features that are useful for measuring lexical semantic change.

In order to reduce the number of model parameters, we do not cluster the resulting word-use representations, but instead directly compare all representations of one relevant word $W_{t_1}$ at time $t_1$ and $W_{t_2}$ at time $t_2$ using a Mean Pairwise Euclidean (MPE) distance metric:

\[
	d_{MPE}(W_{t_1}, W_{t_2}) = \frac{1}{|W_{t_1}| \cdot |W_{t_2}|} \cdot \sum_{w_{t_1} \in W_{t_1}} \sum_{w_{t_2} \in W_{t_2}} \lVert w_{t_1} - w_{t_2} \rVert
\]

\subsection{CIRCE}

We propose the ensemble model CIRCE to combine the predictions of context-free and context-dependent models. The CIRCE rank prediction $r_{CIRCE}$ is generated from the context-free rank $r_{CF}$ and the context-dependent rank $r_{CD}$ through linear combination with a single parameter $\theta \in \mathopen[0.0, 1.0\mathclose]$:

\[
	r_{CIRCE} = \theta \cdot r_{CD} + (1 - \theta) \cdot r_{CF}
\]

Development experiments confirmed that there often are values for $\theta$ that cause CIRCE to perform better than both the context-free and the context-dependent model. This indicates that both kinds of representations contain unique time-specific features that can be exploited in order to predict the degree of lexical semantic change. 

We employ a simple heuristic to calculate $\theta$ in an unsupervised setting such as SemEval-2020 Task 1. Development experiments suggested that the usefulness of context-dependent representations as measured by the optimal value for $\theta_{optimal} = \argmax_{\theta \in \mathopen[0.0, 1.0\mathclose]} \rho(r_{CIRCE}, r_{true})$ roughly correlates with the time classification accuracy of the BERT model after finetuning. Consequently, we predict $\theta$ at test time with a simple linear mapping from the set of realistic BERT classification accuracies $acc_{BERT} \in \mathopen[0.5, 1.0\mathclose]$ to the set of valid CIRCE weights $\theta \in \mathopen[0.0, 1.0\mathclose]$:

\[
	\theta_{CIRCE} = 2 \cdot (acc_{BERT} - 0.5)
\]

\newpage

\section{Experimental Setup}
\label{sec:experimental-setup}

\subsection{Preprocessing}

In the context-free model, we follow \newcite{Schlechtweg2019Woc} in removing words with frequencies below a threshold. We use $\frac{|S|}{5 \cdot 10^{4}} $ for this threshold, where $|S|$ is the number of sentences in the corpus. In order to preserve information, we skip this step for datasets with fewer than $10^{6}$ sentences in total. 

In the context-dependent model, we create a balanced binary time classification dataset from the two diachronic corpora. We employ a  train-test split of $0.8$ / $0.2$ for classification finetuning and evaluation. Furthermore, we create a version of the classification dataset in which words that occur only in one corpus are replaced by the \texttt{[MASK]}-token. The intention behind this preprocessing step is to avoid the learning of rule-based features that are not useful for LSCD such as the memorization of unique words.

\subsection{Implementations}

We use the Word2Vec\footnote{\url{https://github.com/danielfrg/word2vec}} SGNS \cite{Mikolov2013SGNS} implementation to create word vectors and VecMap\footnote{\url{https://github.com/artetxem/vecmap}} \cite{Artetxe2018VecMap} to align them. We use the Transformers\footnote{\url{https://github.com/huggingface/transformers}}  library by HuggingFace \cite{Wolf2019HuggingFacesTS} to finetune BERT and to extract context-dependent word representations.

\subsection{Parameters}

In the context-free model, we create SGNS vectors of dimensionality 300 with window size 10 and negative sample 1. Following \newcite{Schlechtweg2019Woc}, we length-normalize and mean-center the representations when applying orthogonal Procrustes analysis. 

In the context-dependent model, we use the pretrained BERT model \texttt{bert-base-german-cased} in development experiments and \texttt{bert-base-multilingual-cased} in the SemEval submission experiments. We finetune the model with a sequence classification head for one epoch at a learning rate of $4 \cdot 10^{-5}$ with a warm-up step ratio of $0.05$. We extract a context-dependent representation vector of size 768 for each word-use by feeding the whole sentence and taking the mean over the corresponding tokens in the activations of the last hidden layer of BERT.

\section{Results}
\label{sec:results}

\subsection{LSCD Performance}

\begin{table}[h!]
	\begin{tabular}{@{}llp{0.20\textwidth}*{3}{L{\dimexpr0.25\textwidth-2\tabcolsep\relax}}@{}}
		\toprule
		& & Context-Free & CIRCE & Context-Dependent \\
		\midrule
		\multirow{2}{*}{Development} & de-durel & 0.7263 & \underline{0.8018} & 0.4157 \\
		& de-surel & 0.5802 & \underline{0.7251} & 0.7089 \\
		\midrule
		\multirow{4}{*}{Submission} & en-semeval & \underline{0.4221} & 0.2465 & 0.0816 \\
		& de-semeval & \underline{0.7253} & \underline{0.7253} & 0.2075 \\
		& ln-semeval & 0.4124 & \underline{0.4637} & 0.4439 \\
		& sw-semeval & \underline{0.5467} & \underline{0.5467} & 0.0416 \\
		\bottomrule
	\end{tabular}
	\caption{Evaluation phase results for Subtask 2 in development and submission experiments in Spearman's $\uprho$. For each dataset all results with the maximum value are underlined.}
	\label{tab:lscd-results-table}
\end{table}

As Table \ref{tab:lscd-results-table} shows, the context-free model reliably scores well on all datasets for Subtask 2, while the context-dependent model only punctually matches its performance. In cases where both models achieve good results (e.g. de-surel, ln-semeval), the ensemble model CIRCE is able to exceed both predictions.

In the evaluation phase results for Subtask 2, the context-free submission ranks first with a mean correlation of 0.527 over all submission datasets, while the CIRCE submission ranks fifth at a mean correlation of 0.495. The context-dependent submission ranks 69th at a mean correlation of 0.194. 

In the evaluation phase results for Subtask 1, the binarized CIRCE predictions ranks fifth as well with a mean accuracy of 0.639. On en-semeval, it achieved an accuracy of 0.568, on de-semeval 0.728, on ln-semeval 0.550 and on sw-semeval 0.710.

\subsection{Time Classification Performance}

\begin{table}[h!]
	
	\begin{tabular}{@{}lp{0.12\textwidth}*{5}{L{\dimexpr0.19\textwidth-4\tabcolsep\relax}}@{}}
		\toprule
		& \multicolumn{1}{c}{Time Clf. Accuracy} & \multicolumn{2}{c}{CIRCE Weight $\theta$} & \multicolumn{2}{c}{LSCD Spearman's $\uprho$ } \\
		\cmidrule(r{4pt}){2-2} \cmidrule(r{4pt}){3-4} \cmidrule(l){5-6}
		& & $\theta_{CIRCE}$ & $\theta_{optimal}$ & $\theta_{CIRCE}$ & $\theta_{optimal}$ \\
		\midrule
		de-durel & 0.59 & 0.18 & 0.34 & 0.8018 & 0.8772 \\			
		de-surel & 0.95 & 0.90 & 0.92 & 0.7251 & 0.7264 \\			
		en-semeval & 0.82 & 0.64 & 0.08 & 0.2465 & 0.4274 \\
		de-semeval & 0.50 & 0.00 & 0.08 & 0.7253 & 0.7345 \\
		ln-semeval & 0.73 & 0.46 & 0.74 & 0.4637 & 0.4986 \\
		sw-semeval & 0.50 & 0.00 & 0.16 & 0.5467 & 0.5726 \\
		\bottomrule		
	\end{tabular}
	\caption{Time classification accuracies (independent of $\theta$) as well as evaluation phase ($\theta_{CIRCE}$) and post-evaluation phase ($\theta_{optimal}$) CIRCE weights and Subtask 2 performances in Spearman's $\uprho$.}
	\label{tab:time-clf-results-table}
\end{table}

As Table \ref{tab:time-clf-results-table} shows, time classification accuracy of BERT after finetuning varies widely across different datasets. Most remarkably, BERT entirely fails to optimize the time classification objective on de-semeval and sw-semeval, while it achieves exceptionally good results on de-surel.

In general, the classification accuracy seems to be a good predictor for the optimal weight - the submission weight $\theta_{CIRCE}$ calculated with the linear mapping is often within $\pm 0.20$ of the optimal weight $\theta_{optimal}$, and consequently the submission correlations are often close to the optimal correlations.

One notable exception is en-semeval. The high classification accuracy leads to a predicted $\theta_{CIRCE}$ of 0.64, while $\theta_{optimal}$ is located at 0.08. As a result, CIRCE falls short of its potential in the results for Subtask 2 and ranks below the context-free model despite gains on ln-semeval, where CIRCE improves by $0.05$ upon the context-free and by $0.02$ upon the context-dependent model.

\subsection{Error Analysis}

The failure to achieve a significant time classification accuracy for de-semeval and sw-semeval is curious, but might simply be due to the properties of the datasets. Predicting the time period of a given sentence can be a  difficult task even for humans, depending on several factors such as the occurrence of unique words (which are masked in the classification dataset) and the distinctiveness of grammatical structures.

The mismatch of predicted submission weight $\theta_{CIRCE}$ and optimal weight $\theta_{optimal}$ in the case of en-semeval is a more severe error, since it challenges the notion that classification accuracy is a good indicator for representation usefulness. Without extensive experiments, it is difficult to determine the cause of this outlier. However, one contributing factor might be the masking preprocessing step. While the distribution of labels in the BERT classification dataset is balanced, the distribution of \texttt{[MASK]}-tokens is not, since there is often one time period that has more words unique to it than the other. This makes it possible for BERT to learn \texttt{[MASK]}-specific rather than time-specific features during finetuning, which would cause the representations to be useful for time classification but not for LSCD. 

In line with this, repeating the submission experiments without masking in the post-evaluation phase causes the classification accuracy on en-semeval to drop and the predicted submission weight to come within $\pm 0.20$ of the optimal weight for all datasets. This modification boosts the overall mean correlation of CIRCE on the submission datasets to 0.545, which exceeds the performance of all systems for Subtask 2 in the evaluation phase. However, without further experiments on other datasets, the results of this modified model cannot be considered conclusive.

\section{Conclusion}
\label{sec:conclusion}

We presented a context-free model, a context-dependent model and an ensemble model for SemEval-2020 Task 1 Subtask 2. We showed that while the context-free model outperformed all other systems during the evaluation phase, ensembling its predictions with those of the context-dependent model leads to increased performance on the development datasets (de-surel, de-durel) and one submission dataset (ln-semeval) at the cost of decreased performance on another submission dataset (en-semeval).

The submission experiments have made it clear that, although its performance in the development experiments was competitive, the context-dependent model on its own is not a reliable predictor of lexical semantic change. However, it is worth pointing out that the context-dependent representations in all cases contain at least some information that is useful for LSCD, since the optimal CIRCE weight is greater than zero in all experiments. Still, the marginal improvements accomplished with this information do not justify the significant computational effort of finetuning BERT for one epoch.

In further research, it would be interesting to validate the methods described in this paper on additional datasets. In particular, it could be worthwhile to empirically explore the link between classification accuracy and representation usefulness. If the relation were to hold up for other domains and models with a lower computational complexity than BERT, representations obtained through self-supervised classification training could be used on a whole range of other unsupervised tasks.

\bibliographystyle{coling}
\bibliography{circe}

\begin{thebibliography}{}

\bibitem[\protect\citename{Artetxe \bgroup et al.\egroup
  }2018]{Artetxe2018VecMap}
Mikel Artetxe, Gorka Labaka, and Eneko Agirre.
\newblock 2018.
\newblock A robust self-learning method for fully unsupervised cross-lingual
  mappings of word embeddings.
\newblock In {\em Proceedings of the 56th Annual Meeting of the Association for
  Computational Linguistics}, pages 789--798, Melbourne, Australia. Association
  for Computational Linguistics.

\bibitem[\protect\citename{Devlin \bgroup et al.\egroup }2019]{Devlin2019Bert}
Jacob Devlin, Ming-Wei Chang, Kenton Lee, and Kristina Toutanova.
\newblock 2019.
\newblock {BERT}: Pre-training of deep bidirectional transformers for language
  understanding.
\newblock In {\em Proceedings of the 2019 Conference of the North {A}merican
  Chapter of the Association for Computational Linguistics: Human Language
  Technologies, Volume 1 (Long and Short Papers)}, pages 4171--4186,
  Minneapolis, Minnesota. Association for Computational Linguistics.

\bibitem[\protect\citename{Giulianelli}2019]{Giulianelli2019ContextChange}
Mario Giulianelli.
\newblock 2019.
\newblock Lexical semantic change analysis with contextualised word
  representations.
\newblock Master's thesis, University of Amsterdam, Amsterdam, Netherlands.

\bibitem[\protect\citename{Hamilton \bgroup et al.\egroup
  }2016]{Hamilton2016Alignment}
William~L. Hamilton, Jure Leskovec, and Dan Jurafsky.
\newblock 2016.
\newblock Diachronic word embeddings reveal statistical laws of semantic
  change.
\newblock In {\em Proceedings of the 54th Annual Meeting of the Association for
  Computational Linguistics (Volume 1: Long Papers)}, pages 1489--1501, Berlin,
  Germany. Association for Computational Linguistics.

\bibitem[\protect\citename{Han and Eisenstein}2019]{Han2019Finetuning}
Xiaochuang Han and Jacob Eisenstein.
\newblock 2019.
\newblock Unsupervised domain adaptation of contextualized embeddings for
  sequence labeling.
\newblock In {\em Proceedings of the 2019 Conference on Empirical Methods in
  Natural Language Processing and the 9th International Joint Conference on
  Natural Language Processing (EMNLP-IJCNLP)}, pages 4238--4248, Hong Kong,
  China. Association for Computational Linguistics.

\bibitem[\protect\citename{H{\"a}tty \bgroup et al.\egroup
  }2019]{Hatty2019Surel}
Anna H{\"a}tty, Dominik Schlechtweg, and Sabine Schulte~im Walde.
\newblock 2019.
\newblock {SUR}el: A gold standard for incorporating meaning shifts into term
  extraction.
\newblock In {\em Proceedings of the Eighth Joint Conference on Lexical and
  Computational Semantics (*{SEM} 2019)}, pages 1--8, Minneapolis, Minnesota.
  Association for Computational Linguistics.

\bibitem[\protect\citename{Hu \bgroup et al.\egroup }2019]{Hu2019ContextSense}
Renfen Hu, Shen Li, and Shichen Liang.
\newblock 2019.
\newblock Diachronic sense modeling with deep contextualized word embeddings:
  An ecological view.
\newblock In {\em Proceedings of the 57th Annual Meeting of the Association for
  Computational Linguistics}, pages 3899--3908, Florence, Italy. Association
  for Computational Linguistics.

\bibitem[\protect\citename{Kim \bgroup et al.\egroup }2014]{Kim2014StaticTemp}
Yoon Kim, Yi-I Chiu, Kentaro Hanaki, Darshan Hegde, and Slav Petrov.
\newblock 2014.
\newblock Temporal analysis of language through neural language models.
\newblock In {\em Proceedings of the {ACL} 2014 Workshop on Language
  Technologies and Computational Social Science}, pages 61--65, Baltimore, MD,
  USA. Association for Computational Linguistics.

\bibitem[\protect\citename{Liu \bgroup et al.\egroup }2019]{Liu2019ContextRep}
Nelson~F. Liu, Matt Gardner, Yonatan Belinkov, Matthew~E. Peters, and Noah~A.
  Smith.
\newblock 2019.
\newblock Linguistic knowledge and transferability of contextual
  representations.
\newblock In {\em Proceedings of the 2019 Conference of the North {A}merican
  Chapter of the Association for Computational Linguistics: Human Language
  Technologies, Volume 1 (Long and Short Papers)}, pages 1073--1094,
  Minneapolis, Minnesota. Association for Computational Linguistics.

\bibitem[\protect\citename{Mikolov \bgroup et al.\egroup
  }2013a]{Mikolov2013SkipGram}
Tomas Mikolov, Kai Chen, Greg Corrado, and Jeffrey Dean.
\newblock 2013a.
\newblock Efficient estimation of word representations in vector space.

\bibitem[\protect\citename{Mikolov \bgroup et al.\egroup
  }2013b]{Mikolov2013SGNS}
Tomas Mikolov, Ilya Sutskever, Kai Chen, Greg Corrado, and Jeffrey Dean.
\newblock 2013b.
\newblock Distributed representations of words and phrases and their
  compositionality.

\bibitem[\protect\citename{Peters \bgroup et al.\egroup
  }2018]{Peters2018DeepContext}
Matthew Peters, Mark Neumann, Mohit Iyyer, Matt Gardner, Christopher Clark,
  Kenton Lee, and Luke Zettlemoyer.
\newblock 2018.
\newblock Deep contextualized word representations.
\newblock In {\em Proceedings of the 2018 Conference of the North {A}merican
  Chapter of the Association for Computational Linguistics: Human Language
  Technologies, Volume 1 (Long Papers)}, pages 2227--2237, New Orleans,
  Louisiana. Association for Computational Linguistics.

\bibitem[\protect\citename{Schlechtweg \bgroup et al.\egroup
  }2018]{Schlechtweg2018Durel}
Dominik Schlechtweg, Sabine Schulte~im Walde, and Stefanie Eckmann.
\newblock 2018.
\newblock Diachronic usage relatedness ({DUR}el): A framework for the
  annotation of lexical semantic change.
\newblock In {\em Proceedings of the 2018 Conference of the North {A}merican
  Chapter of the Association for Computational Linguistics: Human Language
  Technologies, Volume 2 (Short Papers)}, pages 169--174, New Orleans,
  Louisiana. Association for Computational Linguistics.

\bibitem[\protect\citename{Schlechtweg \bgroup et al.\egroup
  }2019]{Schlechtweg2019Woc}
Dominik Schlechtweg, Anna H{\"a}tty, Marco Del~Tredici, and Sabine Schulte~im
  Walde.
\newblock 2019.
\newblock A wind of change: Detecting and evaluating lexical semantic change
  across times and domains.
\newblock In {\em Proceedings of the 57th Annual Meeting of the Association for
  Computational Linguistics}, pages 732--746, Florence, Italy. Association for
  Computational Linguistics.

\bibitem[\protect\citename{Schlechtweg \bgroup et al.\egroup
  }2020]{Schlechtweg2020TaskDescript}
Dominik Schlechtweg, Barbara McGillivray, Simon Hengchen, Haim Dubossarsky, and
  Nina Tahmasebi.
\newblock 2020.
\newblock {S}em{E}val-2020 {T}ask 1: {U}nsupervised {L}exical {S}emantic
  {C}hange {D}etection.
\newblock In {\em Proceedings of the 14th International Workshop on Semantic
  Evaluation}, Barcelona, Spain. Association for Computational Linguistics.

\bibitem[\protect\citename{Sch{\"o}nemann}1966]{Schonemann1966OrthogonalProcrustes}
Peter Sch{\"o}nemann.
\newblock 1966.
\newblock A generalized solution of the orthogonal procrustes problem.
\newblock {\em Psychometrika}, 31(1):1--10.

\bibitem[\protect\citename{Tahmasebi \bgroup et al.\egroup
  }2018]{Tahmasebi2018Survey}
Nina Tahmasebi, Lars Borin, and Adam Jatowt.
\newblock 2018.
\newblock Survey of computational approaches to lexical semantic change.

\bibitem[\protect\citename{Wolf \bgroup et al.\egroup
  }2019]{Wolf2019HuggingFacesTS}
Thomas Wolf, Lysandre Debut, Victor Sanh, Julien Chaumond, Clement Delangue,
  Anthony Moi, Pierric Cistac, Tim Rault, R'emi Louf, Morgan Funtowicz, and
  Jamie Brew.
\newblock 2019.
\newblock Huggingface's transformers: State-of-the-art natural language
  processing.

\end{thebibliography}

\end{document}